\documentclass{article}

     \usepackage[nonatbib,preprint]{neurips_2020}

\usepackage[utf8]{inputenc} 
\usepackage[T1]{fontenc}    
\usepackage{hyperref}       
\usepackage{url}            
\usepackage{booktabs}       
\usepackage{amsfonts}       
\usepackage{nicefrac}       
\usepackage{microtype}      
\usepackage{subcaption}
\usepackage{xcolor}

\usepackage{graphicx}
\usepackage{color}
\usepackage{comment}
\usepackage{amsmath}
\usepackage{amssymb} 
\usepackage{multirow}

\title{Representing Point Clouds with Generative Conditional Invertible Flow Networks}

\author{%
    Michał Stypułkowski$^{1,2,\dagger}$\\
    \And  
    Kacper Kania$^3$\\
    \And
    Maciej Zamorski$^{2,3}$\\
    \And
    Maciej Zięba$^{2,3}$ \\
    \And
    Tomasz Trzciński$^{2,4}$ \\
    \And
    Jan Chorowski$^{1,5}$ \\
    \and
    
    $^{1}$\text{University of Wrocław} \\
    $^{2}$\text{Tooploox} \\
    $^{3}$\text{Wrocław University of Science and Technology} \\
    $^{4}$\text{Warsaw University of Technology} \\
    $^{5}$\text{NavAlgo} \\
    $^{\dagger}$\texttt{michal.stypulkowski@cs.uni.wroc.pl}
}

\begin{document}

\maketitle

\begin{abstract}
In this paper, we propose a simple yet effective method to represent point clouds as sets of samples drawn from a cloud-specific probability distribution. This interpretation matches intrinsic characteristics of point clouds: the number of points and their ordering within a cloud is not important as all points are drawn from the proximity of the object boundary. 
We postulate to represent each cloud as a \emph{parameterized probability distribution} defined by a generative neural network. Once trained, such a model provides a natural framework for point cloud manipulation operations, such as aligning a new cloud into a default spatial orientation.
To exploit similarities between same-class objects and to improve model performance, we turn to weight sharing: networks that model densities of points belonging to objects in the same family share all parameters with the exception of a small, object-specific embedding vector. We show that these embedding vectors capture semantic relationships between objects.
Our method leverages generative invertible flow networks to learn embeddings as well as to generate point clouds. Thanks to this formulation and contrary to similar approaches, we are able to train our model in an end-to-end fashion. 
As a result, our model offers competitive or superior quantitative results on benchmark datasets, while enabling unprecedented capabilities to perform cloud manipulation tasks, such as point cloud registration and regeneration, by a generative network. 




\end{abstract}

\begin{figure}[!t]
    \centering
    \begin{subfigure}[t]{0.47\linewidth}
        \centering
        \includegraphics[width=\textwidth]{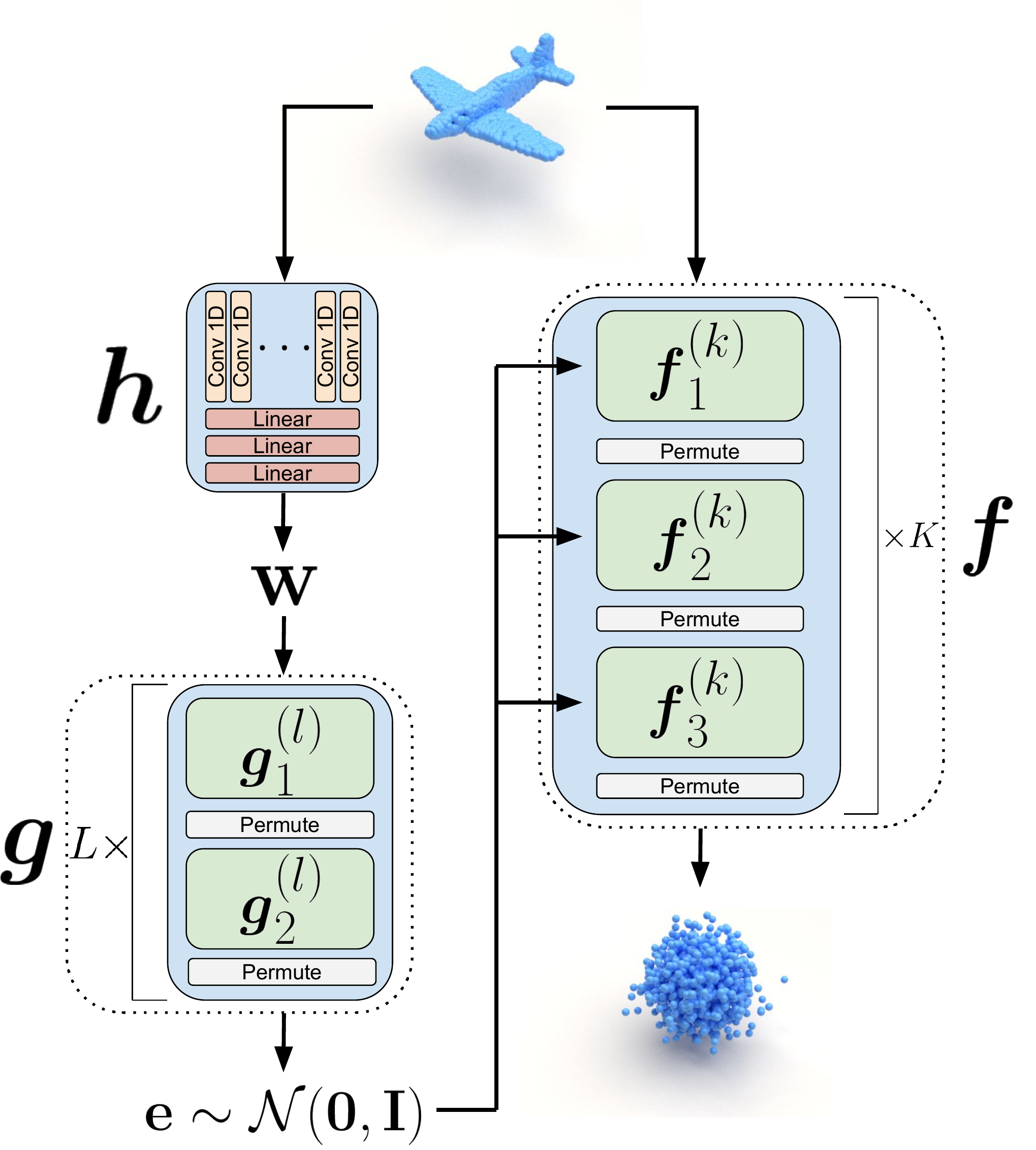}
    \end{subfigure}
    \hfill{}
    \begin{subfigure}[t]{0.47\linewidth}
        \centering
        \includegraphics[width=\textwidth]{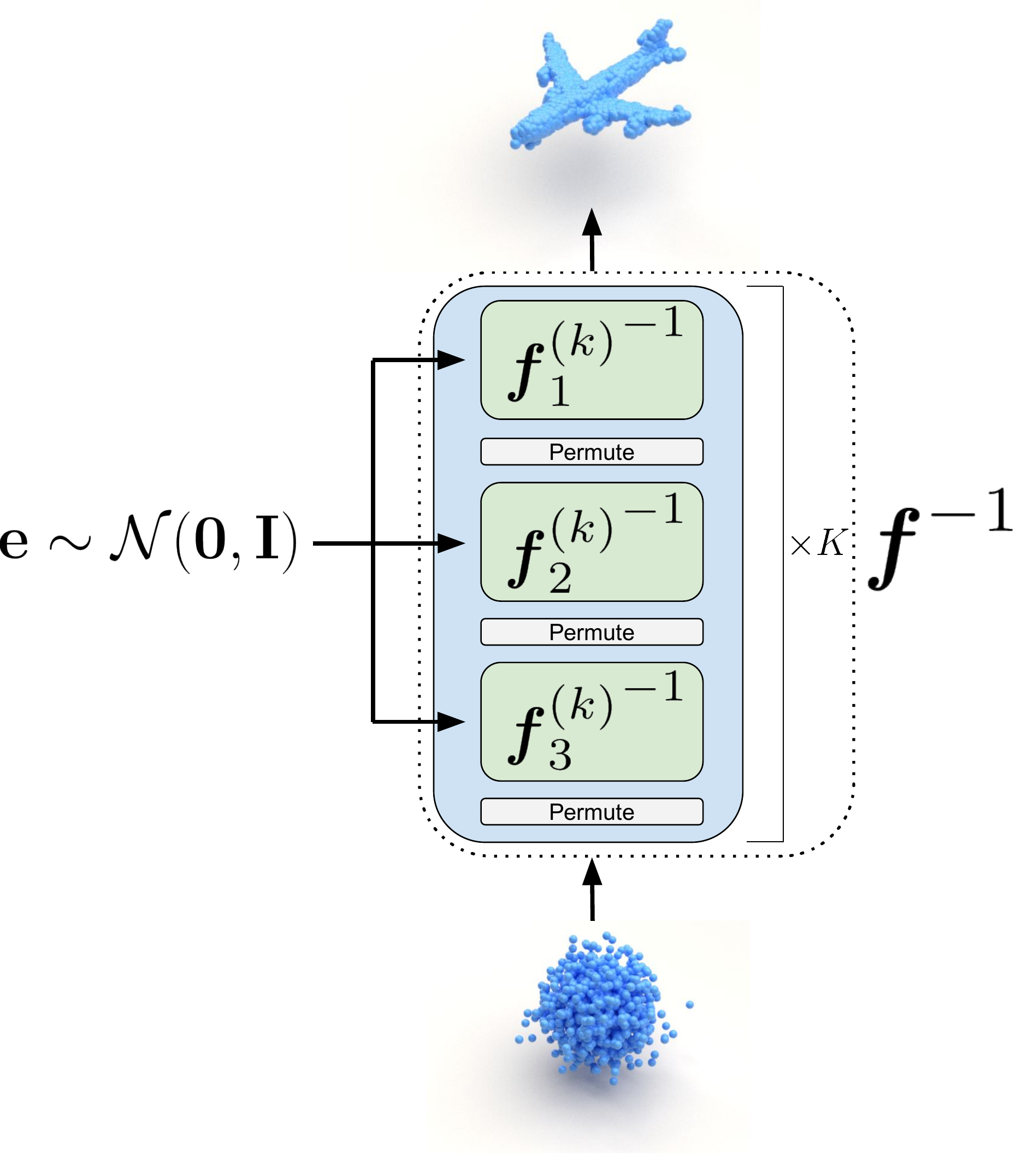}
    \end{subfigure}
    
    \caption{\textbf{Architectures} of our Conditional Invertible Flow Networks. For the inference (left), we first extract initial embedding with encoder $\mathbf{h}$ which is then normalized by flow $\mathbf{g}$. Flow $\mathbf{f}$ normalizes given point cloud conditioned on its embedding. For the sampling (right), we sample points from normal distribution that are transformed by $\mathbf{f}^{-1}$ into shape represented by sampled embedding.
}
  \label{fig:model}
\end{figure}

\section{Introduction}



Proliferation of 3D capturing devices, such as LIDARs and depth cameras gave rise to an unprecedented interest of the research community in processing the so-called \emph{point clouds} - unordered sets of points in three-dimensional space that are created by sampling object surfaces~\cite{wu20153d,qi2016pointnet,qi2017pointnetplusplus,yang2019pointflow,zamorskiadversarial,spurek2020hypernetwork}. This body of research becomes even more relevant nowadays, given a multitude of applications involving 3D space capturing devices, such as robotic manipulation~\cite{kehoe2015survey} and autonomous driving~\cite{yang2018pixor}.

An advent of deep learning applications in 3D point cloud processing started with the introduction of the ShapeNet dataset~\cite{wu20153d}. 
Although first works represented objects and shapes using other 3D representations such as meshes or voxels, point clouds were quickly adopted as a standard. 
Methods that were introduced to process point clouds revolved around discriminative \cite{qi2016pointnet,qi2017pointnetplusplus} and generative modeling \cite{xiang2018generating,xiangyu2018lidar} in point clouds representation.
These models found many applications for graph processing \cite{wang2019graph}, object classification, object detection \cite{qi2018objectdetection,yang2019objectdetection} or segmentation \cite{wu2017squeezeset,boulch2027unstructuredpointcloud}.

In recent years, Variational Autoencoders (VAEs) \cite{kingma2013auto} and Generative Adversarial Networks (GANs) \cite{goodfellow2014generative} arose as one of the most popular frameworks in modeling distribution underlying the input data. In both cases, their application in point clouds often involves optimizing one metrics that measure the discrepancy between true point cloud distribution and prediction. The most well-known are chamfer distance (CD) and earth mover's distance (EMD). However, optimizing these metrics directly \cite{achlioptas2017learning} in GANs can lead to drifting from the true data distribution. On the other hand, VAEs base on an approximation of that distribution through evidence lower bound (ELBO) maximization. Therefore, different approaches were proposed, such as normalizing flows \cite{rezende2015variational}. One of the main advantages of these approaches is the possibility of calculating the likelihood directly. 

The application of deep generative models to 3D shapes was initially studied by \cite{wu20153d} and \cite{wu2016learning}. The former work focuses on the application of the Convolutional Deep Belief Network on 3D point grids. 
The latter uses adversarial training to learn the underlying 3D voxels distribution. 
In \cite{achlioptas2017learning}, the authors provide a two-stage approach with a simple auto-encoder and GAN model trained in stacked mode. 
\cite{zamorskiadversarial} provide an interesting variation of adversarial autoencoder for generating 3D point clouds. 
\cite{li2018point} provided the end-to-end GAN for point clouds. 
One of the most recent methods, PointFlow~\cite{yang2019pointflow}, utilizes VAE, together with continuous normalizing flows \cite{chen2018ode}, to generate 3D point clouds. By conditioning point clouds on shape distribution, PointFlow controls the final shape of the point cloud, as well as facilitates generating novel clouds of high fidelity. 
The distribution is learned through a separate continuous normalizing flow where the prior is Gaussian.

In this paper we propose a novel method for representing and generating 3D point clouds that uses normalized flow-based models. Conceptually, we treat each point cloud as a probability distribution in 3D space for which we train a cloud-specific generative flow-based neural network. This makes our approach inherently order-invariant and insensitive to the different numbers of points among point clouds. 
To capture similarities between point clouds, networks associated with different clouds share parameters. In fact each cloud  provides only a small  individual embedding that is used to condition the shared flow-based model. 
Fig.~\ref{fig:model} shows the overview of our solution. 


The proposed model can be interpreted as a meta-learner, where each point cloud is treated as a separate training set that is used to train a single network. However, instead of training a separate model for each point cloud we use a shared model that takes an additional embedding which controls the shape of the point cloud. Harnessing meta-learning techniques enable us to discover the embedding for new point cloud with gradient-based approach by keeping the parameters of the flow-based model unchanged. 

We identify the following contributions of our work:
\begin{itemize}
    \item a novel generative model for 3D point clouds, 
    \item point-level generative approach that controls the sampling process at every step, is insensitive to the ordering and different numbers of the points, and does not require long training,
    \item a new meta-learning approach, where each point cloud (as separate training subset) is represented by its unique embedding. 
\end{itemize}

\section{Model}

The architecture of our model is presented in Figure \ref{fig:model}. We treat individual point clouds as probability distributions in $\mathbb{R}^3$. The normalizing flow network $\mathbf{f}$ implements an invertible random variable transformation between the standard normal prior and the points in the cloud. The flow $\mathbf{f}$ is conditioned on latent vectors $\mathbf{e}$ that are also normally distributed. We compute the conditioning vectors using a PointNet encoder $\mathbf{h}$ whose output, $\mathbf{w}$ is transformed to the standard normal prior using another normalizing flow $\mathbf{g}$. The whole model is trained jointly by direct optimization of a single loss function.  


To generate new object-specific points clouds, we first sample a cloud embedding $\mathbf{e}$ from the standard normal prior. Then, we use the sampled embedding $\mathbf{e}$ to condition the model $\mathbf{f}$. Finally, we generate individual points in the cloud by sampling from the standard normal and transforming them into the data space using $\mathbf{f}^{-1}$. 

\subsection{Conditional Flow-based Cloud Generator} \label{sec:flow_model}
The main idea of flow-based models is to find an invertible transformation between the unknown data distributions $p_X$ and known simple prior distribution $p_Z$.  
Hence, the goal of flow-based models is to transforms $p_X$ to match a simpler, tractable distribution $p_Z$ which is defined manually when constructing the model.  We assume that there exists a function $\mathbf{f}$ such that $\mathbf{z}=\mathbf{f}(\mathbf{x}, \mathbf{e})$ and $\mathbf{x}=\mathbf{f}^{-1}(\mathbf{z}, \mathbf{e})$, where $\mathbf{z}\sim p_Z$, $\mathbf{x}\sim p_X$, and $\mathbf{e}$ is an arbitrary conditioning. We make use of the change of variables formula for density functions:
\begin{align}
    p_X(\mathbf{x} | \mathbf{e}) = p_Z(\mathbf{f}(\mathbf{x}, \mathbf{e})) \left| \det \frac{\partial \mathbf{f}(\mathbf{x}, \mathbf{e})}{\partial \mathbf{x}} \right|,
\end{align}
where $\left| \det \frac{\partial \mathbf{f}(\mathbf{x}, \mathbf{e})}{\partial \mathbf{x}} \right|$ is the absolute value of the determinant of the Jacobian of the transformation $\mathbf{f}$ at point $\mathbf{x}$. With a proper definition of $\mathbf{f}$, we are able to train the conditional model with direct likelihood of the data. We follow \cite{dinh2016density} in construction of our flows and define $\mathbf{f}$ as a combination of simple flows $\mathbf{f}^{(k)}_i$, each followed by permutation of the dimensions, where $\mathbf{f}^{(k)}_i$ is defined as:
\begin{equation}
    \mathbf{f}^{(k)}_i \bigg{(}
    \begin{bmatrix}
        \mathbf{y}_{1} \\
        \mathbf{y}_{2}
    \end{bmatrix}\bigg{)} = 
    \begin{bmatrix}
        \mathbf{y}_{1} \\
        \mathbf{y}_{2} \odot \exp (M(\mathbf{y}_{1}, \mathbf{e})) + A(\mathbf{y}_{1}, \mathbf{e})
    \end{bmatrix},
    \label{simple_flow}
\end{equation}
where $\mathbf{y}$ is an output from the previous transformation and $\big[\begin{smallmatrix} \mathbf{y}_1\\ \mathbf{y}_2 \end{smallmatrix}\big]$ is some partition of the input. This transformation is easily invertible and the logarithm of the Jacobian equals to $\sum_i M(\mathbf{y}_{1})_i$. Also, since $f = f_n \circ f_{n-1} \circ \dots \circ f_1$, we get $\frac{\partial f}{\partial \mathbf{x}} = \frac{\partial f_1}{\partial \mathbf{x}} \cdot \frac{\partial f_2 }{\partial f_1(\mathbf{x})} \cdot \dots \cdot \frac{\partial f_n}{\partial f_{n-1} \circ \dots \circ f_1(\mathbf{x})}$. Note that there are no constraints on either $M$ or $A$ thus we choose them to be neural networks.

To define our training objective, we sample two subsets of points $\mathbf{x}^f$, $\mathbf{x}^h$ from each cloud in a training batch and train the model by maximizing the likelihood of the points $\mathbf{x}^f$ conditioned on a cloud embedding computed on $\mathbf{x}^h$. First, we put the points in $\mathbf{x}^h$ through a trainable PointNet encoder $\mathbf{h}$ to obtain a cloud representation $\mathbf{w}$ which we then map to the conditioning $\mathbf{e}$ using another normalizing flow network $\mathbf{g}$:
\begin{equation}
    \begin{split}
    \mathbf{w} &= \mathbf{h}(\mathbf{x}^h) \\
    \mathbf{e} &= \mathbf{g}(\mathbf{w}) 
    \end{split}
\end{equation}

Finally, we use points in $\mathbf{x}^f$ to compute the conditional cloud likelihood:
\begin{equation}
    \begin{split}
    \log p_X(\mathbf{x}^f|\mathbf{x}^h) &= \sum_{i=1}^N \log p_Z(\mathbf{f}(\mathbf{x}_i^f,e)) + \log \left| \det \frac{\partial \mathbf{f}(\mathbf{x}^f_i, \mathbf{e})}{\partial \mathbf{x}^f_i} \right| 
    \end{split}
\end{equation}

We extend our loss function with a regularization term which promotes a normal distribution of the conditioning vectors $\mathbf{e}$:
\begin{equation}
    \begin{split}
    \log p_W(\mathbf{w}|\mathbf{x}^h) &= \log p_E(\mathbf{g}(\mathbf{w})|X) + \log \left| \det \frac{\partial \mathbf{g}(\mathbf{e})}{\partial \mathbf{e}} \right| 
    \end{split}
\end{equation}

Our final loss function is defined to be:
\begin{equation}
    \begin{split}
    \mathcal{L}=-\log p_X(\mathbf{x}^f|\mathbf{x}^h) - \log p_W(\mathbf{w}|\mathbf{x}^h).
    \end{split}
\end{equation}

PointFlow \cite{yang2019pointflow} pioneered the use of normalizing flows to represent point clouds. However, it uses a continuous flow, which takes long time to train, while we employ a simpler discrete flow. Our parameterization of the decoder allows easy injection of domain knowledge, such as separate control of object shape and position. Moreover, PointFlow is a variational auto-encoder: it feeds a set of points to the encoder, then reconstructs them and relies on the VAE stochastic encoder with reparameterization trick to prevent learning an identity encoding. To achieve similar regularization effect without using VAE, we train the model to reconstruct points different than the ones used by the encoder. This also prevents trivial identity data representations, but is deterministic and does not require sampling. Taken jointly, these changes lead to fast and stable training convergence.

\section{Experiments}

\begin{figure*}
  \centering
  \includegraphics[width=\textwidth]{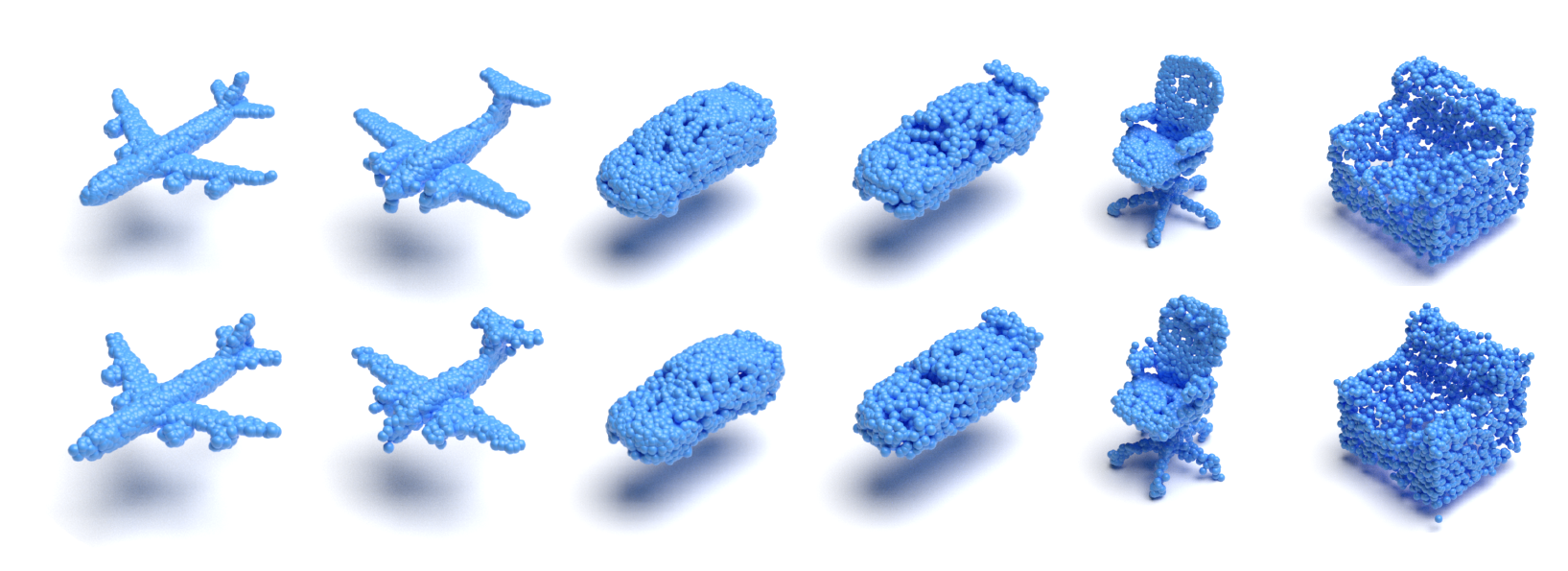}
  \caption{\textbf{Reconstructions} of our model obtained with an autoencoding architecture presented in~Figure~\ref{fig:recon-model}. The first row shows inputs to the model, and the second - their reconstructions.}
  \label{fig:recon}
\end{figure*}

The goal of the experiments is to provide quantitative and qualitative analysis of reconstructive and generative capabilities of the model in terms of the quality of reconstructed objects, generated samples, interpolation and metrics including coverage and minimum matching distance. For all experiments we used point clouds from ShapeNet dataset preserving the validation split from \cite{yang2019pointflow}. 

\subsection{Architecture}
We implemented networks $\mathbf{f}$ and $\mathbf{g}$ as normalizing flows. Network $\mathbf{f}$ is a combination of $10$ segments $\mathbf{f}^{(k)}$ for $k=1,...,10$ with $3$ blocks $\mathbf{f}^{(k)}_i$ for $i=1,2,3$ each. Every block is defined by two ResNets. Analogically $\mathbf{g}$ has $5$ segments with $2$ blocks each.
The encoder $\mathbf{h}$ follows architecture designs of PointNet (\cite{qi2016pointnet}).
It consists of 1D convolutional layers followed by features extraction and fully connected layers.
We chose dimensions of $\mathbf{w}$ and $\mathbf{e}$ to equal $32$. Priors $p_Z$ and $p_E$ are $d$-dimensional standard normal distributions, where $d$ equals to $3$ and $32$ respectively. We used Adam optimizer with default parameters and learning rate $10^{-4}$ decaying every $10$ epochs by factor $0.8$.


\subsection{Results}


\begin{figure*}
  \centering
  \includegraphics[width=\textwidth]{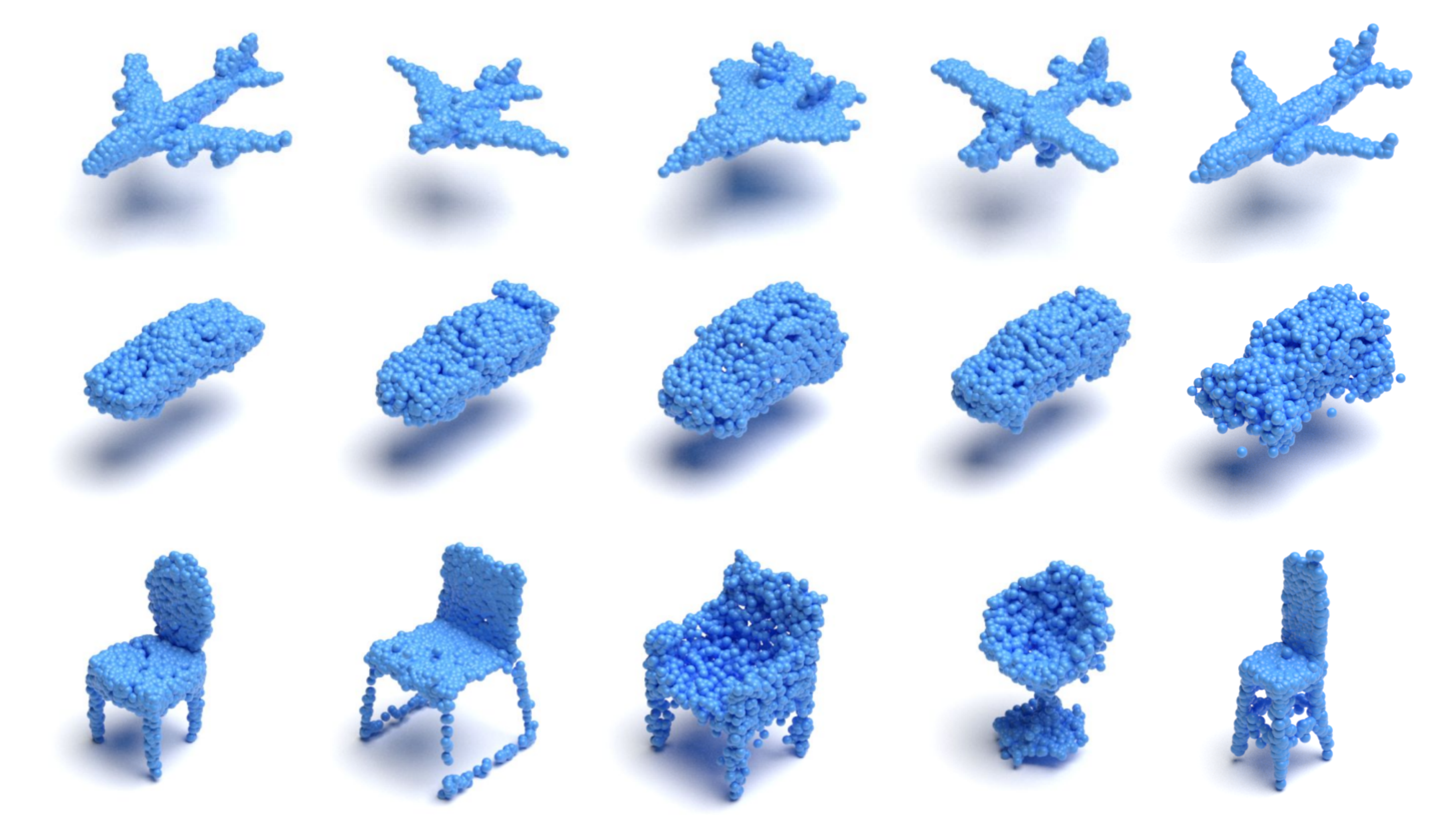}
  \caption{\textbf{Samples} generated with $\mathbf{f}$ normalizing flow. Rows show samples from different categories: airplanes, cars and chairs.}
  \label{fig:samples}
\end{figure*}

\begin{figure}
\centering
\begin{minipage}[c]{0.47\linewidth}
    \centering
    \includegraphics[width=0.8\textwidth]{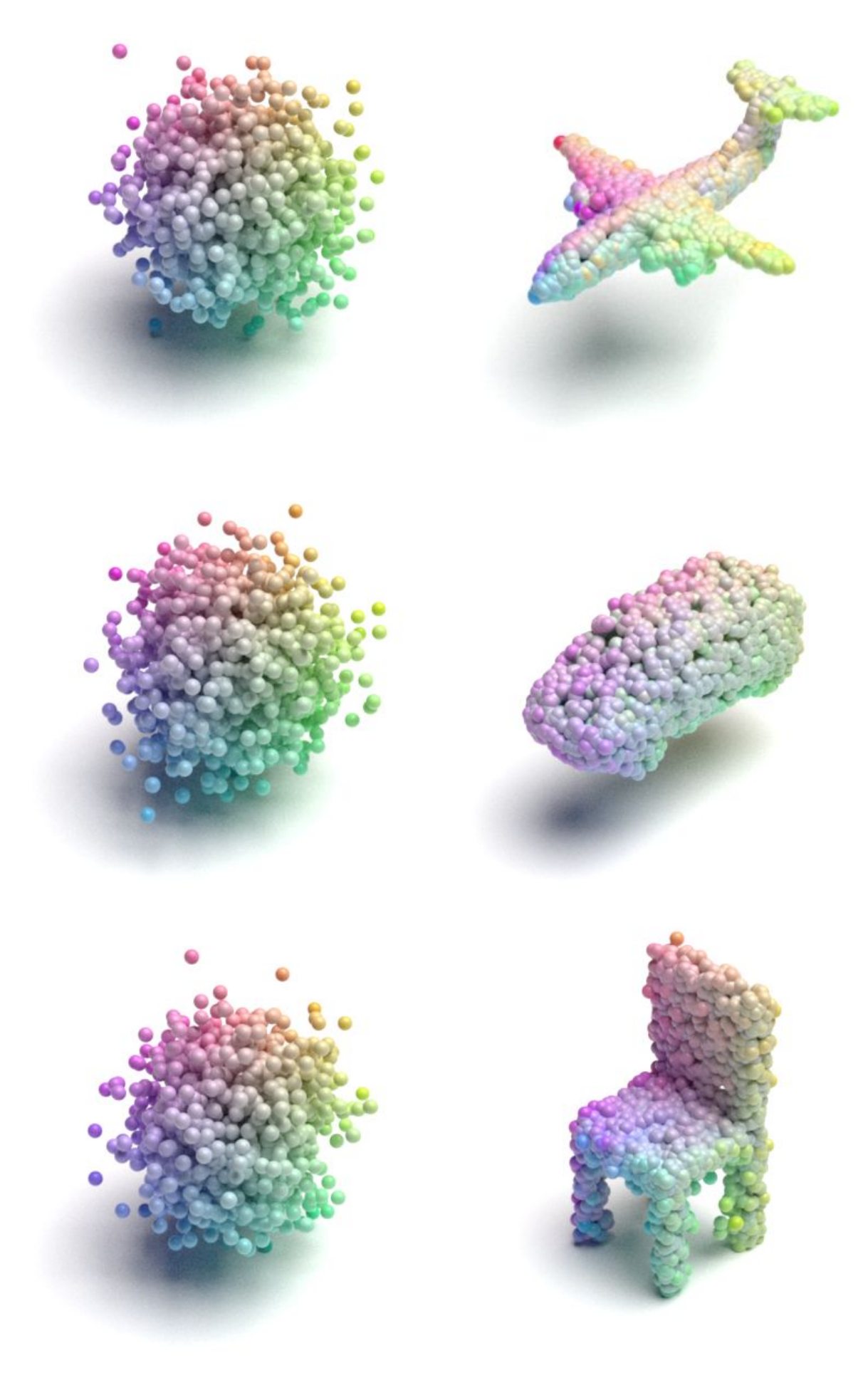} \quad
\end{minipage}
\hfill{}
\begin{minipage}[c]{0.47\linewidth}
    \centering
    \includegraphics[width=\textwidth]{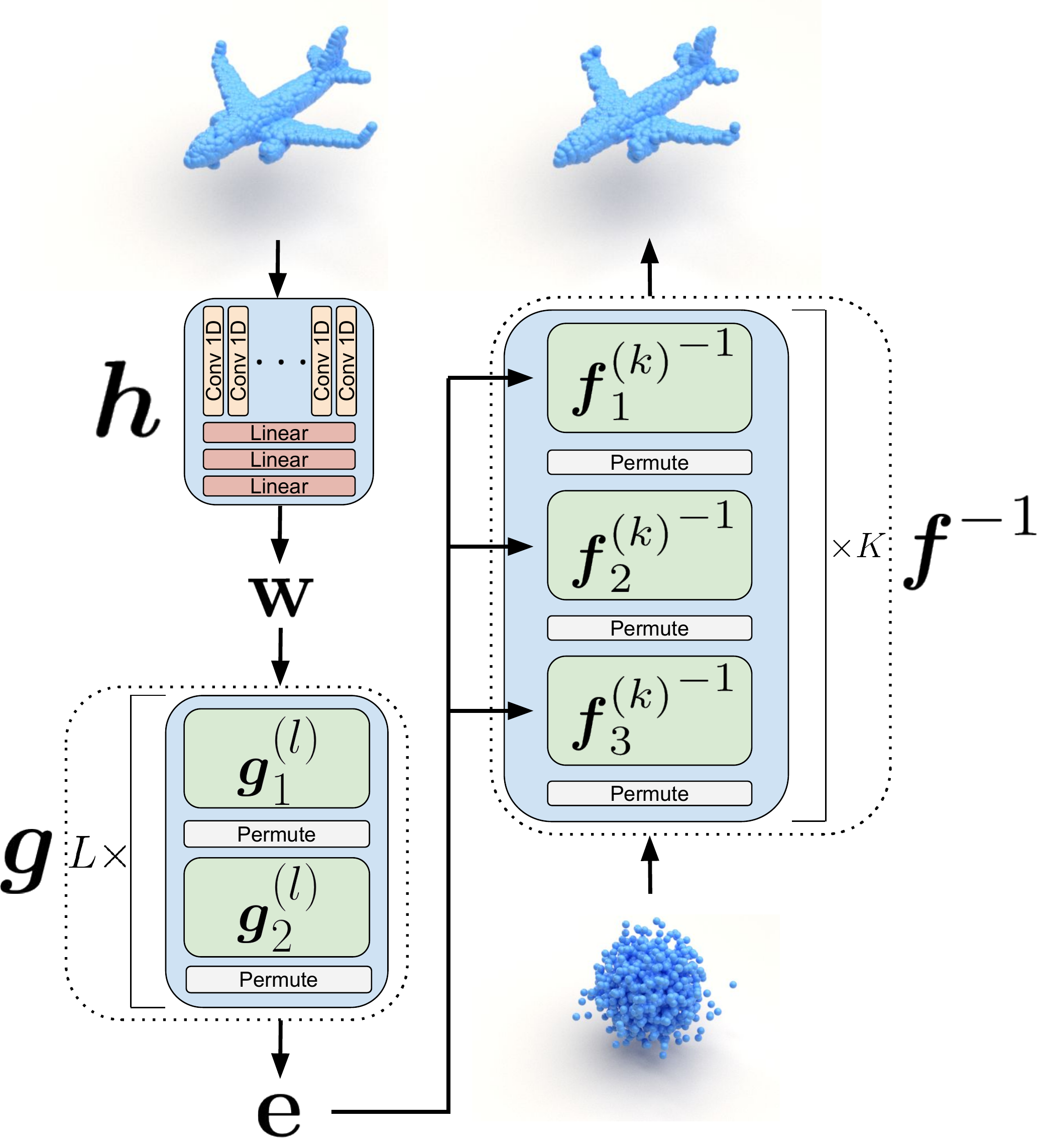}
\end{minipage}

\begin{minipage}[t]{0.47\linewidth}
    \caption{Point clouds sampled from $\mathcal{N}(0, I)$ (left) and reconstructed using our model (right). Note, how the model maps points to the closest semantic parts of objects.}
    \label{fig:recon-samples}
\end{minipage}
\hfill{}
\begin{minipage}[t]{0.47\linewidth}
    \caption{Model used during for the reconstruction task. Since we leverage invertible normalizing flows, the $\mathbf{f}$ modules can serve as a decoder of an autoencoder architecture.}
    \label{fig:recon-model}
\end{minipage}
\end{figure}

\paragraph{Reconstruction} As the first experiment, we test the reconstruction capabilities of our model according to the procedure outlined in Figure \ref{fig:recon-model}. 

First, we pass an input sample $\mathbf{x}$ through the encoding network $\mathbf{h}$ to obtain the feature vector $\mathbf{w}$, which is further used as an input to the flow $\mathbf{g}$. As a result, we obtain an $\mathbf{e}$ which is embedding representing given cloud $\mathbf{x}$.

In order to reconstruct the object, we pass the points sampled from the $p_Z \sim \mathcal{N}(0, I)$ and the conditioning latent code $\mathbf{e}$ as the input to our inverted flow $\mathbf{f}^{-1}$ to produce the reconstructed point cloud.

The results are presented in the Figure~\ref{fig:recon}. They show ability of our model to correctly encode unseen objects into the latent space and then reconstruct them using a desired number of points.


\paragraph{Sampling}
To check our model's ability to generalize to the whole data space, we perform a sampling experiment. 
In order to sample a new point cloud, we first sample an embedding $\mathbf{e}$ from our prior distribution $p_E$. 
With that, we obtain the conditioning term for a network $\mathbf{f}$. 
We then sample a chosen number of points $\mathbf{z}$ from a distribution $p_Z$.
The embedding $\mathbf{e}$ and points $\mathbf{z}$ are then used as an input to $\mathbf{f}^{-1}$ in order to generate various shapes of point clouds.
Figure \ref{fig:samples} presents the qualitative results of the sampling experiment.

Following generation procedure in \cite{kingma2018glow}, we found that for cars and chairs, sampling $\mathbf{e}$ from the prior distribution with increased standard deviation results in better samples both in terms of quality and metrics performance. We used a standard deviation equal to $1.25$ and $1.3$ for cars and chairs, respectively.

\begin{figure*}
  \centering
  \includegraphics[width=\textwidth, trim={0 0.58cm 0 0.3cm}, clip]{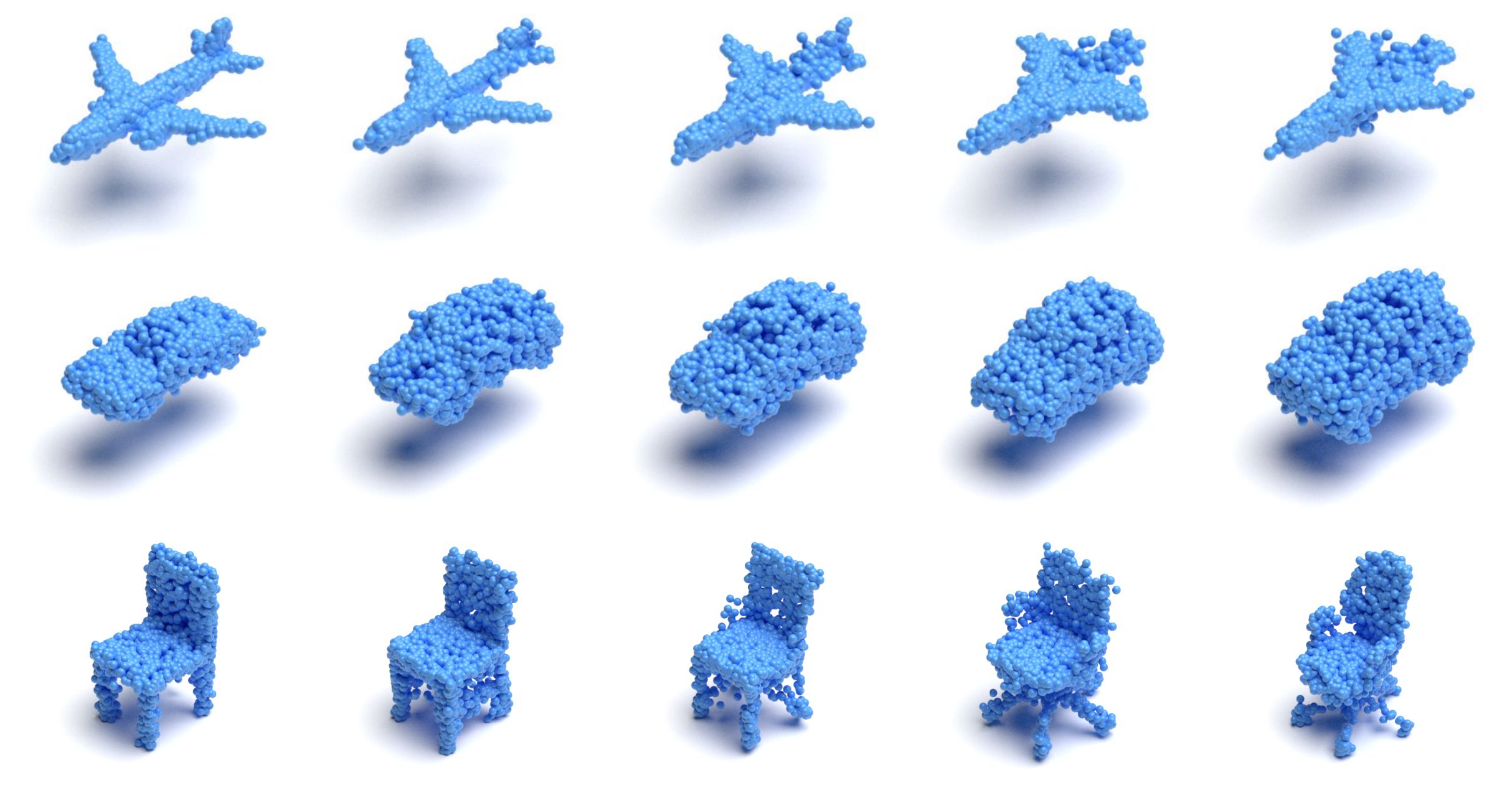}
  \caption{\textbf{Interpolation} of latent vectors $\mathbf{e}$ makes smooth transition between their reconstructions (from left to right). Each row shows interpolation on different classes of shapes.}
  \label{fig:inter}
\end{figure*}

\paragraph{Interpolation}
As a way to assess the embedding space $E$ continuity, we perform the linear interpolation between the true point cloud shapes. 
First, we take two samples and obtain their latent space embeddings. 
Next, we calculate the intermediate latent codes between them. 
We then use those new latent space codes as an input to the $\mathbf{f}^{-1}$ to obtain interpolated point clouds. 
Results presented in Figure \ref{fig:inter} show that embeddings are arranged naturally and logically. 
Transitions between each sample are smooth, and there are no substantial visual differences.

\begin{figure}
    \centering
    \includegraphics[width=\textwidth]{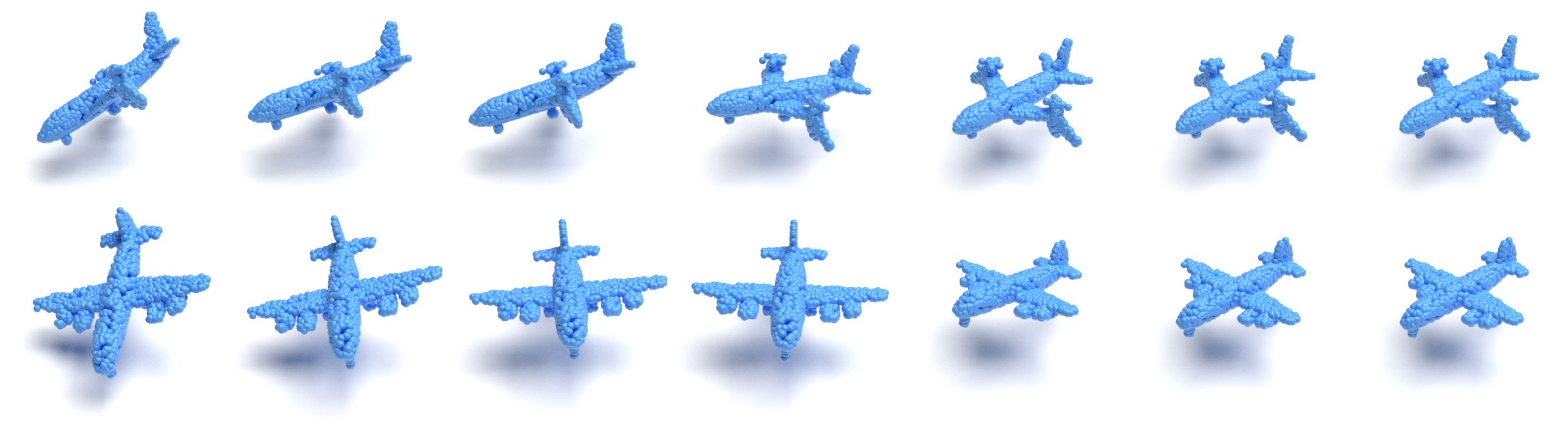}
    \caption{CIF is able to align objects into a canonical position. We extend the flow $\mathbf{f}$ by a multiplication by a 3D rotation matrix. We then optimize the cloud likelihood when conditioned on a null embedding vector with respect to this matrix. Each row shows subsequent steps of the optimization, starting from a random pose. The last image in each row is the object in the training set. We see that CIF is able to recover the default object pose used in the training data.}
    \label{fig:rotate_into_pos}
\end{figure}

\paragraph{Object alignment} 3D point transformations, such as rotations or scalings can be modeled as specialized normalizing flow layers and added to the flow network $\mathbf{f}$. In Figure \ref{fig:rotate_into_pos} we show the possibility of recovering object pose. We extend the flow $\mathbf{f}$ with multiplication by a parametric rotation matrix. We then randomly rotate a cloud and optimize its conditional likelihood $p_X(\mathbf{x}|\mathbf{e}=0)$ with respect to the rotation matrix.  During the procedure, we use the null embedding vector $\mathbf{e}=0)$ to deprive the model of cloud shape information. For optimization we have used the CMA-ES procedure \cite{hansen_cma}. We can see that this procedure is able to recover the default object pose used in the training dataset (pictured in the last column). 

\begin{figure*}
  \centering
    \begin{subfigure}[t]{0.47\linewidth}
        \centering
        \includegraphics[width=\textwidth]{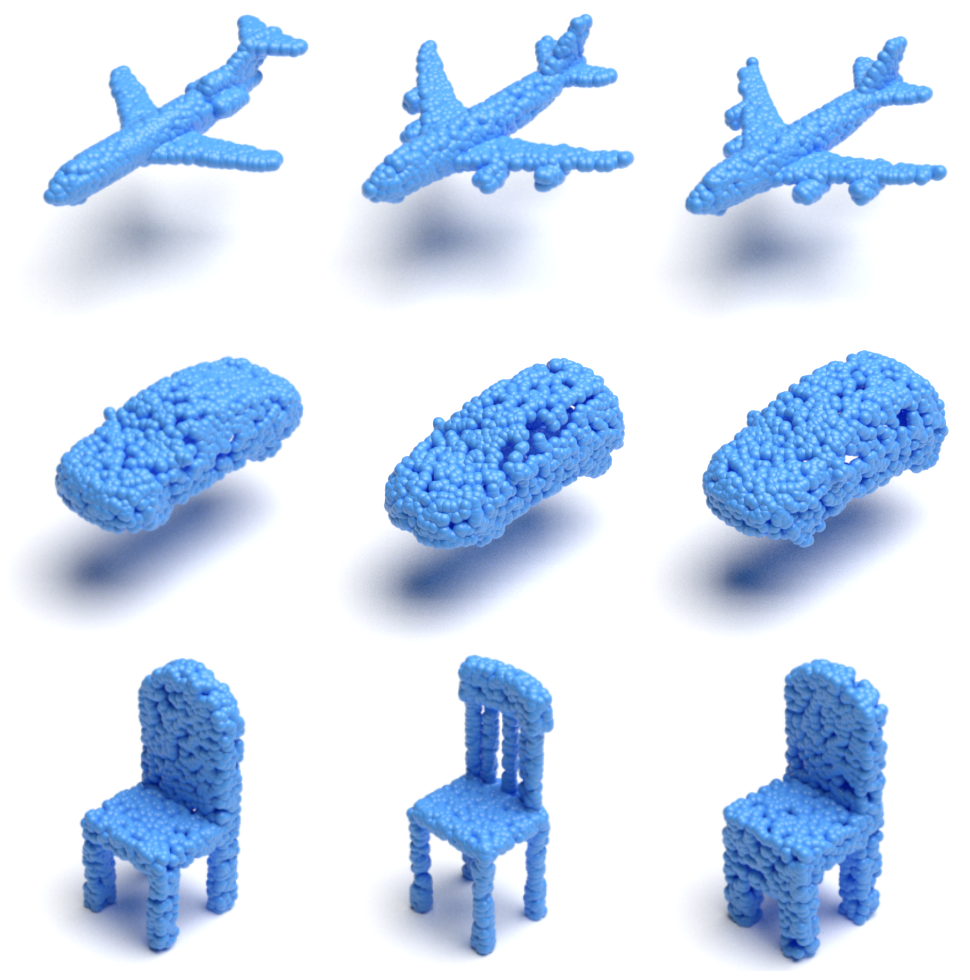}
    \end{subfigure}
    \hfill{}
    \begin{subfigure}[t]{0.47\linewidth}
        \centering
        \includegraphics[width=\textwidth]{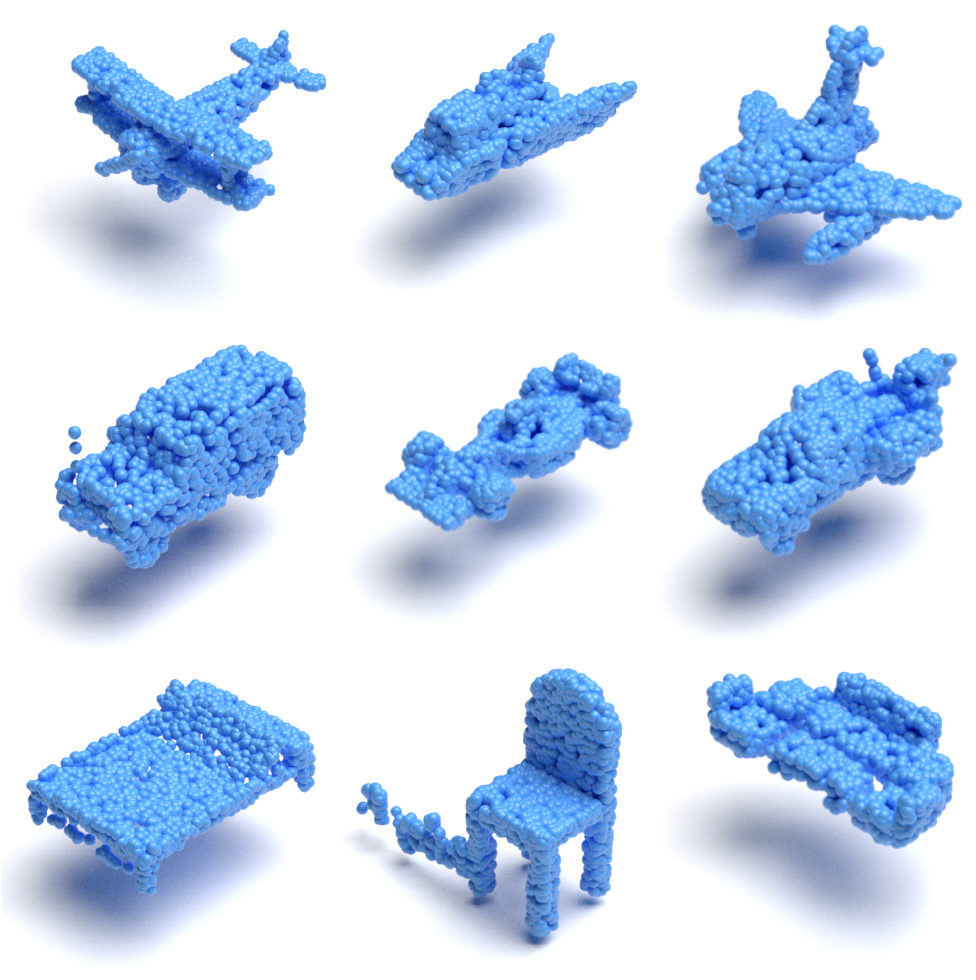}
    \end{subfigure}
  \caption{CIF is able to find the common (leftmost three clouds in each row) and rare (rightmost three) samples in the dataset by comparing the prior probability of their embeddings. Notice that using this approach, we detected erroneous samples in the dataset.}
  \label{fig:likelihood}
\end{figure*}

\paragraph{Determining typical objects and outliers} We show that one can evaluate the likelihood $p_W(\mathbf{w}|\mathbf{x}^h)$ of the embedding vector of a given point cloud, which can be further used to determine the uniqueness of samples. We process point clouds from the dataset through $\mathbf{h}$ encoder and pass it to the $\mathbf{g}$ flow. Then, we calculate the negative likelihood of the embedding being sampled from the prior distribution $\mathcal{N}(0, I)$. The "rarest" samples in the dataset, according to the model, can be found by choosing point clouds with the top highest negative likelihood. These samples can be further investigated in outlier detection procedure. Similarly, we can find the most typical point clouds by selecting samples producing the top lowest negative log-likelihoods. Using this method, we show in Figure \ref{fig:likelihood} unique and typical samples from the ShapeNet dataset.



\begin{table*}[]
\begin{center}
\caption{Generation results. MMD-CD scores are multiplied by $10^3$; MMD-EMD scores are multiplied by $10^2$; JSDs are multiplied by $10^2$.}
\begin{tabular}{llccccccc}
\toprule
\multirow{2}{*}[-2pt]{Category} & \multirow{2}{*}[-2pt]{Methods} & \multirow{2}{*}[-2pt]{JSD $(\downarrow)$} & \multicolumn{2}{c}{MMD $(\downarrow)$} & \multicolumn{2}{c}{COV $(\uparrow)$} & \multicolumn{2}{c}{1-NNA $(\downarrow)$} \\ \cmidrule{4-9} 
                          &                          &                      & CD          & EMD       & CD         & EMD        & CD          & EMD         \\ \midrule
\multirow{7}{*}{Airplane}  & r-GAN                    & 7.44                 & 0.261       & 5.47      & 42.72      & 18.02      & 93.58       & 99.51       \\
                          & l-GAN (CD)               & 4.62                 & 0.239       & 4.27      & 43.21      & 21.23      & 86.30       & 97.28       \\
                          & l-GAN (EMD)              & \textbf{3.61}                 & 0.269       & 3.29      & \textbf{47.90}      & 50.62      & 87.65       & 85.68       \\
                          & PC-GAN                   & 4.63                 & 0.287       & 3.57      & 36.46      & 40.94      & 94.35       & 92.32       \\
                          & PointFlow                & 4.92                 & \textbf{0.217}       & 3.24      & 46.91      & 48.40      & \textbf{75.68}       & 75.06       \\
                          & CIF (ours)               & 4.24                 & 0.221       & \textbf{3.14}      & 47.57      & \textbf{52.67}      & 77.08       & \textbf{72.59}       \\ \cmidrule{2-9} 
                          & Training set             & \textcolor{gray}{6.61}                 & \textcolor{gray}{0.226}       & 3.08      & \textcolor{gray}{42.72}      & \textcolor{gray}{49.14}      & 70.62      & 67.53       \\ \midrule
\multirow{7}{*}{Chair}     & r-GAN                    & 11.5                 & 2.57        & 12.8      & 33.99      & 9.97       & 71.75       & 99.47       \\
                          & l-GAN (CD)               & 4.59                 & 2.46        & 8.91      & 41.39      & 25.68      & 64.43       & 85.27       \\
                          & l-GAN (EMD)              & 2.27                 & 2.61        & 7.85      & 40.79      & 41.69      & 64.73       & 65.56       \\
                          & PC-GAN                   & 3.90                 & 2.75        & 8.20      & 36.50      & 38.98      & 76.03       & 78.37       \\
                          & PointFlow                & 1.74                 & 2.42        & 7.87      & \textbf{46.83}      & 46.98      & \textbf{60.88}       & \textbf{59.89}       \\
                          & CIF                      & \textbf{1.42 }       & \textbf{2.38}    & \textbf{7.85}  & 44.01      & \textbf{47.03}      & 62.71       & 63.39             \\ \cmidrule{2-9} 
                          & Training set             & \textcolor{gray}{1.50}                 & 1.92        & 7.38      & 57.25      & 55.44      & 59.67       & 58.46       \\ \midrule
\multirow{7}{*}{Car}       & r-GAN                    & 12.8                 & 1.27        & 8.74      & 15.06      & 9.38       & 97.87       & 99.86       \\
                          & l-GAN (CD)               & 4.43                 & 1.55        & 6.25      & 38.64      & 18.47      & 63.07       & 88.07       \\
                          & l-GAN (EMD)              & 2.21                 & 1.48        & 5.43      & 39.20      & 39.77      & 69.74       & 68.32       \\
                          & PC-GAN                   & 5.85                 & 1.12        & 5.83      & 23.56      & 30.29      & 92.19       & 90.87       \\
                          & PointFlow                & 0.87                 & 0.91        & 5.22      & 44.03      & 46.59      & \textbf{60.65}       & 62.36       \\
                          & CIF                      & \textbf{0.79}        & \textbf{0.90} & \textbf{5.12} & \textbf{44.79} & \textbf{49.24}  & 64.82 & \textbf{61.36}   \\ \cmidrule{2-9} 
                          & Training set                & \textcolor{gray}{0.86}                 & \textcolor{gray}{1.03}        & \textcolor{gray}{5.33}      & 48.30      & 51.42      & 57.39       & 53.27       \\ \bottomrule
\end{tabular}
\label{tab:gen_results}
\end{center}
\end{table*}
\subsection{Quantitative analysis}
Following the methodology proposed in \cite{yang2019pointflow} we evaluate the generative capabilities of the model with the criteria: Jensen-Shannon Divergence (JSD), Coverage (COV), Minimum Matching Distance (MMD) and 1-Nearest Neighbor Accuracy (1-NNA). For last three of the mentioned measures we consider using Chamfer (COV-CD, MMD-CD, 1-NNA-CD) and Earth-Mover's (COV-EMD, MMD-EMD, 1-NNA-EMD) distances. 

 We compare the results with the existing solutions including: raw-GAN \cite{achlioptas2017learning}, latent-GAN \cite{achlioptas2017learning}, PC-GAN \cite{li2018point} and PointFlow \cite{yang2019pointflow}. We train each model using point clouds from one of the three categories in the ShapeNet dataset: \emph{airplane},
\emph{chair}, and \emph{car}. We sample 2048 points for each of the point cloud used for evaluation. We also report the performance of evaluation metrics for point clouds from the training set, which is considered as an upper bound since they are from the target distribution. The results of the quantitative analysis are presented in Table \ref{tab:gen_results}. The CIF model achieves a competitive results comparing to the reference approaches. Moreover, our approach does not require extensive training procedure, essential during application of ODE solver for continuous flows utilized by PointFlow. More precisely, results reported in Table \ref{tab:gen_results} were achieved after 4 days of training on 2 RTX 2080 Ti GPUs, while PointFlow was trained for 2 weeks on 8 V100s.


\section{Conclusion}
We demonstrate a novel point-level, order-invariant method to encode and generate point clouds. The model treats point clouds as probability distributions and represents them using normalized flow-based models. It can be trained end-to-end efficiently, and be applied for generating novel 3D point clouds that follow governing data distribution. We evaluated the performance of the method in both a qualitative and quantitative manner. Our model sets new state-of-the-art of generation quality on many parts of presented measures and performs on par with reference approaches on the rest.  Moreover, we show that our model can have a variety of applications such as object alignment, typical object finding, and outlier detection.


\section*{Broader Impact}
We believe that our research brings much progress in generative modeling of point clouds. Main benefits can be noticed in applications of LIDAR or depth sensors, where the point cloud representation is common. However, the raw point cloud is difficult to process due to intrinsic noise during data gathering. Moreover, these representations are often difficult to process. Our model provides hidden representations of input point clouds that can be efficiently analyzed. Additionally, the model is sound from the probability theory perspective since it models parametrized distribution, which resembles the original distribution of the data. 

Current generative deep learning approaches suffer from being specialized in modeling a single class of objects. We consider it as a problem that can create potential risks in areas such as autonomous driving, where the versatility of algorithms is of much importance for real-time processing. Thus, industries often opt for classical computer vision to extract information from point clouds. For this reason, we encourage researchers to further improve the generative processing of point clouds to make it beneficial in a variety of applications.



\bibliographystyle{splncs04}
\bibliography{bib}

\begin{thebibliography}{10}
\providecommand{\url}[1]{\texttt{#1}}
\providecommand{\urlprefix}{URL }
\providecommand{\doi}[1]{https://doi.org/#1}

\bibitem{achlioptas2017learning}
Achlioptas, P., Diamanti, O., Mitliagkas, I., Guibas, L.: Learning
  representations and generative models for 3d point clouds. arXiv preprint
  arXiv:1707.02392  (2017)

\bibitem{boulch2027unstructuredpointcloud}
Boulch, A., Saux, B.L., Audebert, N.: {Unstructured Point Cloud Semantic
  Labeling Using Deep Segmentation Networks}. In: Pratikakis, I., Dupont, F.,
  Ovsjanikov, M. (eds.) Eurographics Workshop on 3D Object Retrieval. The
  Eurographics Association (2017). \doi{10.2312/3dor.20171047}

\bibitem{chen2018ode}
Chen, T.Q., Rubanova, Y., Bettencourt, J., Duvenaud, D.: Neural ordinary
  differential equations. CoRR  \textbf{abs/1806.07366} (2018),
  \url{http://arxiv.org/abs/1806.07366}

\bibitem{dinh2016density}
Dinh, L., Sohl-Dickstein, J., Bengio, S.: Density estimation using real nvp.
  arXiv preprint arXiv:1605.08803  (2016)

\bibitem{goodfellow2014generative}
Goodfellow, I., Pouget-Abadie, J., Mirza, M., Xu, B., Warde-Farley, D., Ozair,
  S., Courville, A., Bengio, Y.: Generative adversarial nets. In: Advances in
  neural information processing systems. pp. 2672--2680 (2014)

\bibitem{hansen_cma}
Hansen, N.: The {CMA} evolution strategy: {A} comparing review. In: Lozano,
  J.A., Larra{\~{n}}aga, P., Inza, I., Bengoetxea, E. (eds.) Towards a New
  Evolutionary Computation - Advances in the Estimation of Distribution
  Algorithms, Studies in Fuzziness and Soft Computing, vol.~192, pp. 75--102.
  Springer (2006). \doi{10.1007/3-540-32494-1\_4},
  \url{https://doi.org/10.1007/3-540-32494-1\_4}

\bibitem{kehoe2015survey}
Kehoe, B., Patil, S., Abbeel, P., Goldberg, K.: A survey of research on cloud
  robotics and automation. IEEE Transactions on automation science and
  engineering  \textbf{12}(2),  398--409 (2015)

\bibitem{kingma2013auto}
Kingma, D.P., Welling, M.: Auto-encoding variational bayes. arXiv preprint
  arXiv:1312.6114  (2013)

\bibitem{kingma2018glow}
Kingma, D.P., Dhariwal, P.: Glow: Generative flow with invertible 1x1
  convolutions. In: Advances in Neural Information Processing Systems. pp.
  10215--10224 (2018)

\bibitem{li2018point}
Li, C.L., Zaheer, M., Zhang, Y., Poczos, B., Salakhutdinov, R.: Point cloud
  gan. arXiv preprint arXiv:1810.05795  (2018)

\bibitem{qi2018objectdetection}
Qi, C.R., Liu, W., Wu, C., Su, H., Guibas, L.J.: Frustum pointnets for 3d
  object detection from rgb-d data. In: The IEEE Conference on Computer Vision
  and Pattern Recognition (CVPR) (June 2018)

\bibitem{qi2016pointnet}
Qi, C.R., Su, H., Mo, K., Guibas, L.J.: Pointnet: Deep learning on point sets
  for 3d classification and segmentation. CoRR  \textbf{abs/1612.00593} (2016),
  \url{http://arxiv.org/abs/1612.00593}

\bibitem{qi2017pointnetplusplus}
Qi, C.R., Yi, L., Su, H., Guibas, L.J.: Pointnet++: Deep hierarchical feature
  learning on point sets in a metric space. CoRR  \textbf{abs/1706.02413}
  (2017), \url{http://arxiv.org/abs/1706.02413}

\bibitem{rezende2015variational}
Rezende, D.J., Mohamed, S.: Variational inference with normalizing flows (2015)

\bibitem{spurek2020hypernetwork}
Spurek, P., Winczowski, S., Tabor, J., Zamorski, M., Zi{\k{e}}ba, M.,
  Trzci{\'n}ski, T.: Hypernetwork approach to generating point clouds.
  Proceedings of the 37th International Conference on Machine Learning (ICML)
  (2020)

\bibitem{wang2019graph}
Wang, Y., Sun, Y., Liu, Z., Sarma, S.E., Bronstein, M.M., Solomon, J.M.:
  Dynamic graph cnn for learning on point clouds. ACM Trans. Graph.
  \textbf{38}(5) (Oct 2019). \doi{10.1145/3326362},
  \url{https://doi.org/10.1145/3326362}

\bibitem{wu2017squeezeset}
Wu, B., Wan, A., Yue, X., Keutzer, K.: Squeezeseg: Convolutional neural nets
  with recurrent {CRF} for real-time road-object segmentation from 3d lidar
  point cloud. CoRR  \textbf{abs/1710.07368} (2017),
  \url{http://arxiv.org/abs/1710.07368}

\bibitem{wu2016learning}
Wu, J., Zhang, C., Xue, T., Freeman, B., Tenenbaum, J.: Learning a
  probabilistic latent space of object shapes via 3d generative-adversarial
  modeling. In: Advances in neural information processing systems. pp. 82--90
  (2016)

\bibitem{wu20153d}
Wu, Z., Song, S., Khosla, A., Yu, F., Zhang, L., Tang, X., Xiao, J.: 3d
  shapenets: A deep representation for volumetric shapes. In: Proceedings of
  the IEEE conference on computer vision and pattern recognition. pp.
  1912--1920 (2015)

\bibitem{xiang2018generating}
Xiang, C., Qi, C.R., Li, B.: Generating 3d adversarial point clouds. CoRR
  \textbf{abs/1809.07016} (2018), \url{http://arxiv.org/abs/1809.07016}

\bibitem{yang2018pixor}
Yang, B., Luo, W., Urtasun, R.: Pixor: Real-time 3d object detection from point
  clouds. In: Proceedings of the IEEE conference on Computer Vision and Pattern
  Recognition. pp. 7652--7660 (2018)

\bibitem{yang2019objectdetection}
Yang, B., Luo, W., Urtasun, R.: {PIXOR:} real-time 3d object detection from
  point clouds. CoRR  \textbf{abs/1902.06326} (2019),
  \url{http://arxiv.org/abs/1902.06326}

\bibitem{yang2019pointflow}
Yang, G., Huang, X., Hao, Z., Liu, M.Y., Belongie, S., Hariharan, B.:
  Pointflow: 3d point cloud generation with continuous normalizing flows. arXiv
  preprint arXiv:1906.12320  (2019)

\bibitem{xiangyu2018lidar}
Yue, X., Wu, B., Seshia, S.A., Keutzer, K., Sangiovanni{-}Vincentelli, A.L.: A
  lidar point cloud generator: from a virtual world to autonomous driving. CoRR
   \textbf{abs/1804.00103} (2018), \url{http://arxiv.org/abs/1804.00103}

\bibitem{zamorskiadversarial}
Zamorski, M., Zieba, M., Klukowski, P., Nowak, R., Kurach, K., Stokowiec, W.,
  Trzcinski, T.: Adversarial autoencoders for compact representations of 3d
  point clouds. arXiv preprint arXiv:1811.07605  (2018)

\end{thebibliography}

\end{document}